# Behavior evolution-inspired approach to walking gait reinforcement training for quadruped robots

Yu Wang[1], Wenchuan Jia[1, *], Yi Sun[1], Dong He[1]

[1] School of Mechatronic Engineering and Automation, Shanghai University, Shanghai, China.

[*] Corresponding author: lovvchris@shu.edu.cn

**Abstract**

Reinforcement learning method is extremely competitive in gait generation techniques for quadrupedal robot, which is mainly due to the fact that stochastic exploration in reinforcement training is beneficial to achieve an autonomous gait. Nevertheless, although incremental reinforcement learning is employed to improve training success and movement smoothness by relying on the continuity inherent during limb movements, challenges remain in adapting gait policy to diverse terrain and external disturbance. Inspired by the association between reinforcement learning and the evolution of animal motion behavior, a self-improvement mechanism for reference gait is introduced in this paper to enable incremental learning of action and self-improvement of reference action together to imitate the evolution of animal motion behavior. Further, a new framework for reinforcement training of quadruped gait is proposed. In this framework, genetic algorithm is specifically adopted to perform global probabilistic search for the initial value of the arbitrary foot trajectory to update the reference trajectory with better fitness. Subsequently, the improved reference gait is used for incremental reinforcement learning of gait. The above process is repeatedly and alternatively executed to finally train the gait policy. The analysis considering terrain, model dimensions, and locomotion condition is presented in detail based on simulation, and the results show that the framework is significantly more adaptive to terrain compared to regular incremental reinforcement learning.

**Keywords:** Gait training; Quadruped robot; Reinforcement learning; Behavior evolution; Reference trajectory; Genetic algorithm.

## 1. Background

In recent years, legged robots such as quadruped robots have shown increasingly powerful locomotion capability, which is largely due to the ongoing intensive research and application of modern control techniques such as Model-Predictive Control [1-3]. Nevertheless, we still believe that reinforcement learning is extremely competitive in gait generation. Although the model predictive control method based on the optimal control concept and state prediction enables the robot to solve the joint input condition required to reach the expected state in real time, the expected state of the robot still needs to be planned in advance, and therefore the autonomy of the robot locomotion is not fundamentally improved. Moreover, the model prediction results are closely related to the initial value of the state, and thus are uncertain and difficult to reproduce accurately. In contrast, reinforcement learning theory is based on the effective imitation of the learning process of real behaviors, including gait learning, where fully autonomous gait can be acquired [4] after sufficient training for a reasonable reward target.

Reinforcement learning (RL) method [5] was first applied to discrete system, whose state space, action space, and trained policy are expressed in discrete form, which motivated its debut in TV game tasks. As continuous reinforcement learning methods have been proposed successively [6], not only simple physical systems in classical control can obtain the expected motion, but also more complex continuous tasks such as quadruped gait policy have been trained successfully [4][7]. In particular, reinforcement learning methods such as DDPG [8], A3C [9], PPO [10], and SAC [11] have been demonstrated to be applicable to continuous system.

Due to the significant increase in the state space dimension and action space dimension, reinforcement training for continuous tasks becomes difficult, and the trained movement policy is less stable than real movement. Considering that the action sequences in such systems are usually naturally continuous, incremental reinforcement learning has thus been proposed [12], which employs the continuous characteristics of the system to improve the success rate of training and enhance the stability of the movement. For the quadruped gait training task, incremental reinforcement learning has been a popular approach nowadays [13]. Kim, Shao, and Tan all achieved good results with incremental reinforcement training by introducing the guide mechanism as a reference or a priori [14-16]. Nevertheless, the challenge of achieving a more suitable gait and adapting the quadruped robot to various terrains and resisting external disturbances through reinforcement learning methods remains, which is the motivation for this study.

Since reinforcement learning mechanism is similar to the acquisition process of animal motion behavior, we explore the difference and relevance by comparing them. Then, inspired by their intrinsic relationship, the existing training framework based on reinforcement learning to generate gait is improved to fundamentally enhance the success rate of training and retain the autonomy of gait generation. The detailed results presented in this paper demonstrate the significant effect of this improvement. A comparison of animal motion behavior evolution and reinforcement learning mechanism is shown in Fig. 1, where the former is displayed in the left region and the latter is displayed in the right region, with a specific instance of quadrupedal walking shown in the central region, separated by thick dotted lines between adjacent regions. From behavioral perspective, gait walking is classified as an instinctive behavior, which can be considered as a sequence of reflex behaviors, i.e., the spontaneous generation of continuous movements based on underlying perceptual information from internal and external sources, such as foot state and changes in center of gravity. Instinctive behaviors evolved from reflex behaviors [17-18], the latter usually referring to simple movements evoked by simple external stimulus signals, such as joint-level movements like the knee-jump reflex. This evolution is based on the long-term regulation of movements by factors including body balance, energy consumption, and habituation. The evolved gait is capable of employing the whole body to achieve movement and subsequently undergoes progressive self-improvement so that the gait is continually regulated to better adapt to the growing body and the various nature environment. This process also partly explains why reinforcement learning has difficulties in natural gait learning. For reinforcement learning, the training subject is usually a simplified rigid mechanism, which is significantly different from real musculoskeletal system with good elastic motion, and the factors considered in the design of the reward function are usually only a subset of the many factors that influence the behavior evolution (shown as the dark gray shaded area in Fig. 1), making it difficult to achieve realistic limb movement by relying only on training of joint movement. Incremental reinforcement learning can be understood as deriving from and incorporating progressive feature in behavior evolution, as shown by the red dashed line. Moreover, we believe that it is crucial to introduce a self-improvement mechanism for the reference gait (shown as the red solid line in Fig. 1), where incremental learning and self-improvement for the gait together constitute an imitation of the evolution of the animal's walking behavior. A new framework constructed based on the idea of behavioral evolution for gait reinforcement training is presented in this paper, which incorporates both incremental learning of gait movement and self-improvement capability, and its good performance in gait training of quadruped robot is also demonstrated.

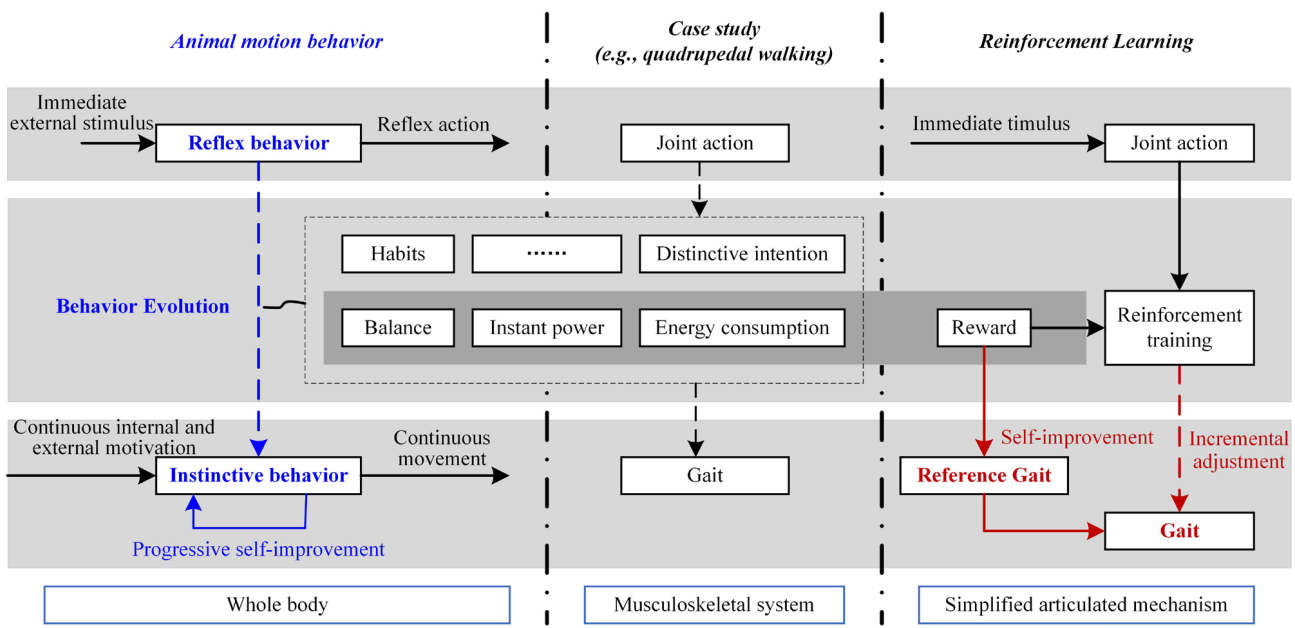

Fig. 1. Improving existing reinforcement training processes based on evolution concept of animal motion behavior.

## 2. Related Works

### 2.1. Problem Statement

Reinforcement learning can usually be described as Markov Decision Process (MDP), while the motion control of a quadruped robot can be considered as a Partially Observable Markov Decision Processes (POMDP) described by the tuple $(S, A, P, R, \gamma)$, where the state space $S$ and the action space $A$ are assumed to be continuous, and the state transition probability $P: S \times S \times A \rightarrow [0,1]$ denotes the probability density of the next state $s_{t+1} \in S$ when the current state $s_t \in S$ and action $a_t \in A$ are set. The environment emits a reward $R: S \times A \rightarrow [R_{min}, R_{max}]$ on each transition, $\gamma \in [0,1]$ represents a discount factor. $\pi(a_t|s_t): S \times A \rightarrow [0,1]$ is the action probability distribution based on the state. The stochastic policy algorithm specifies an action probability distribution at each state, and the deterministic policy algorithm specifies a certain action at each state. And Maximum entropy reinforcement learning optimizes policies $\pi^*$ to maximize both the expected reward and the expected entropy of the policy, as shown in Eq. (1).

$$\pi^* = argmax E_\pi \left[ \sum_t R(s_t, a_t) + \alpha H(\pi(\cdot|s_t)) \right] \quad (1)$$

where $\alpha$ is temperature parameter that determines the relative importance of the entropy term versus the reward, and thus controls the stochasticity of the optimal policy. $H(\pi(\cdot|s_t))$ is the entropy of $\pi$ at state $s_t$, which is calculated as Eq. (2).

$$H(\pi(\cdot|s_t)) = -log\pi(\cdot|s_t) \quad (2)$$

The core idea of maximum entropy is to explore as far as possible without missing any useful action. Compared with deterministic policy that ends the learning process after finding an optimal path, maximum entropy-based reinforcement learning has a stronger exploration capability because it explores various optimal possibilities in different ways to avoid falling into local optima. This means that the learned policy can more easily find better patterns and be more robust under multimodal rewards, and can make more reasonable adjustments when disturbed [19].

### 2.2. Soft Actor-Critic

Challenges of some commonly used deep reinforcement learning algorithms in continuous state and action spaces include: New samples need to be collected at each gradient step, so the number of gradient steps required and the number of samples required at each step both increase with the complexity of the task; the effect of both off-policy learning, and network-based approximation of high-dimensional nonlinear function on stability and convergence; the fragility and sensitivity of the hyperparameters result in the need to use the appropriate parameter values for different tasks.

Soft Actor-Critic (SAC) [11] is a newly proposed off-policy maximum entropy deep reinforcement learning in recent years, which can be applied to complex high-dimensional tasks. It has been shown to perform well and stably in various commonly used Mujoco benchmark tasks (e.g., humanoid walking task with 21 action dimensions, and quadrupedal jumping task with 154 observation dimensions [20]) and real robot control tasks (e.g., Minitaur learns walking gait directly with eight direct-drive actuators, and a 3-finger dexterous robotic hand manipulates an object [21]).

The state value function $V(s_t)$ of SAC can be expressed by Eq. (3).

$$V(s_t) = E_\pi[Q(s_t, a_t) - \alpha log\pi(a_t|s_t)] \quad (3)$$

SAC parameterizes soft Q-function $Q_\theta(s_t, a_t)$ and a tractable policy $\pi_\phi(a_t|s_t)$ and the parameters of networks are $\theta$ and $\emptyset$. The soft Q-function parameters can be trained to minimize the soft Bellman residual by Eqs. (4) to (5).

$$J_Q(\theta) = E_{(s_t,a_t) \sim D} \left[ \frac{1}{2} \left( Q_\theta(s_t, a_t) - \hat{Q}(s_t, a_t) \right)^2 \right] \quad (4)$$

with

$$\hat{Q}(s_t, a_t) = r(s_t, a_t) + \gamma E_{s_{t+1} \sim p}[V_{\bar{\theta}}(s_{t+1})] \quad (5)$$

The policy parameters can be learned by directly minimizing the expected KL-divergence, as shown in Eq. (6).

$$J_\pi(\phi) = E_{s_t \sim D, a_t \sim \pi_\phi} \left[ log\pi_\phi(a_t|s_t) - \frac{1}{\alpha} Q_\theta(s_t, a_t) + logZ(s_t) \right] \quad (6)$$

It is only necessary to keep collecting data and reducing the values of these two loss functions to complete training.

## 2.3. Genetic Algorithm

The dynamic equation of the legged robot is high-order and nonlinear, which requires the gait generation by searching for solutions of the parameters in a very irregular multidimensional space [22], so the standard gradient search-based optimization methods are not well suited for legged gait that with high degrees of freedom [23]. To obtain a natural and efficient gait two strategies can be followed to coordinate the movement of the legs, the first one assuming that the gait of the animals is optimal (otherwise they would not survive in the wild) and thus using the real motion data of the animals to drive the robot to generate the gait. However, the differences in kinematics and dynamics between animals and legged robots make behavior modification necessary to be introduced [24]. The second one is to convert the gait generation of a legged robot into a multi-objective optimization problem containing multiple constraints [25]. Evolutionary algorithm is a global probabilistic search based on biological evolutionary mechanism such as natural selection and genetic variation, and has been shown to be effective in solving such large-scale, multi-objective, multi-constrained optimization problems.

Genetic algorithm (GA) is one of the Evolutionary Computation (EC) solvers and is one of the most popular tools for gait optimization. Chernova and Veloso designed adaptive mechanisms with different mutation and crossover probabilities to optimize the gait of AIBO [22]. Miguel et al. described the quadruped crawling gait as a multi-objective optimization problem by using GA to explore the parameter space of the CPG network to obtain the gait with minimum body vibration, maximum velocity and maximum stability margin [26]. Chae et al. divided the elliptical trajectory of the foot into three phases and searched for the optimal intersection location of each phase to obtain a globally optimized foot trajectory based on energy and stability criteria and using GA [27]. Evolutionary algorithm usually requires the maintenance of a population of candidate solutions and a large number of population iterations, and therefore, evaluating each candidate gait can be very time-consuming. Especially when experimenting on realistic robots, computational efficiency becomes one of the biggest limitations in EC-based gait optimization. To better achieve gait optimization, we believe it is necessary to study EC-based optimization algorithms instead of relying on EC alone for learning and optimization.

## 3. Framework

Although controllers trained by reinforcement learning have advantages that cannot be replaced by manual design, the increase in complexity of the environment and task leads to the creation of a large state space and makes the disadvantage of low sample utilization in traditional reinforcement learning more significant, in addition to more detailed reward settings that increase the difficulty of the problem. In order to achieve spontaneous rhythmic movements of the quadruped robot and accelerate the reinforcement learning process, we propose a new gait reinforcement training framework based on the idea of behavior evolution as shown in Fig. 2. This framework consists of two main parts: the reference action generator (the area in the dashed box) and the reinforcement training process (the light blue area).

The reference trajectory $a_{ref}(t)$ that is generated by the reference action generator can be directly used as an a priori reference to accelerate the training during the reinforcement learning of quadruped gait. In a previous study [28], the reference action generator only produces a set of reference trajectories or actions initially, and the subsequent training tasks are all performed by reinforcement learning. In contrast, inspired by behavior evolution, we design the Reference Trajectory Optimizer inside the Reference Action Generator (as shown in the gray area in the figure) to self-improve the reference action, so that the reference action is constantly adjusted to better suit the environment. The detailed process of the reference action being self-improved is shown by Algorithm 1. This reference action generator requires two types of initial reference information, one is a rhythm signal $\rho$ generated by Central Pattern Generator (CPG) for associating gait phase, and this $\rho$ consists of motion rhythm signals of the hip and knee joints. The Hopf oscillator with a simple form is employed in this paper as the Central Pattern Generator, which can also be replaced by other types of oscillators. Another one is the initial foot trajectory $\varphi_0$, which is related to the rhythm signal $\rho$ using a Radial Basis Function Network (RBFN). RBFN has a simple structure and its local mapping feature can greatly accelerate learning and avoid local minima. The state $R_i$ of the hidden layer neurons in RBFN is represented by Eq. (7).

$$R_i = \exp\left\{-\frac{\Sigma(\rho_j - \mu_{i,j})^2}{\sigma_{RBF}^2}\right\}, i = 1, \dots, H; j = 1,2,3,4 \qquad (7)$$

where $i$ is the index of the hidden layer neurons, $H$ is the total number of RBFN neurons, $\rho$ is the joint position signal,

$j$ is leg index, $\mu_{i,j}$ is the RBFN neuron mean which is calculated by Eq. (8), and $\sigma_{RBF}^2$ is the common variance for the four means.

$$\mu_{i,j} = \rho_j\left(\frac{(i-1)*T}{H-1}\right), j = 1,2,3,4 \tag{8}$$

where $T$ is the period of the CPG signal, and $H = 20$, $\sigma_{RBF}^2 = 0.04$, $T = 2$. The linear layer output of RBFN can be directly used as the reference value for the action of each joint of the quadruped robot, as expressed in Eq. (9).

$$a_{ref}(t) = W * R_i(t) + b \tag{9}$$

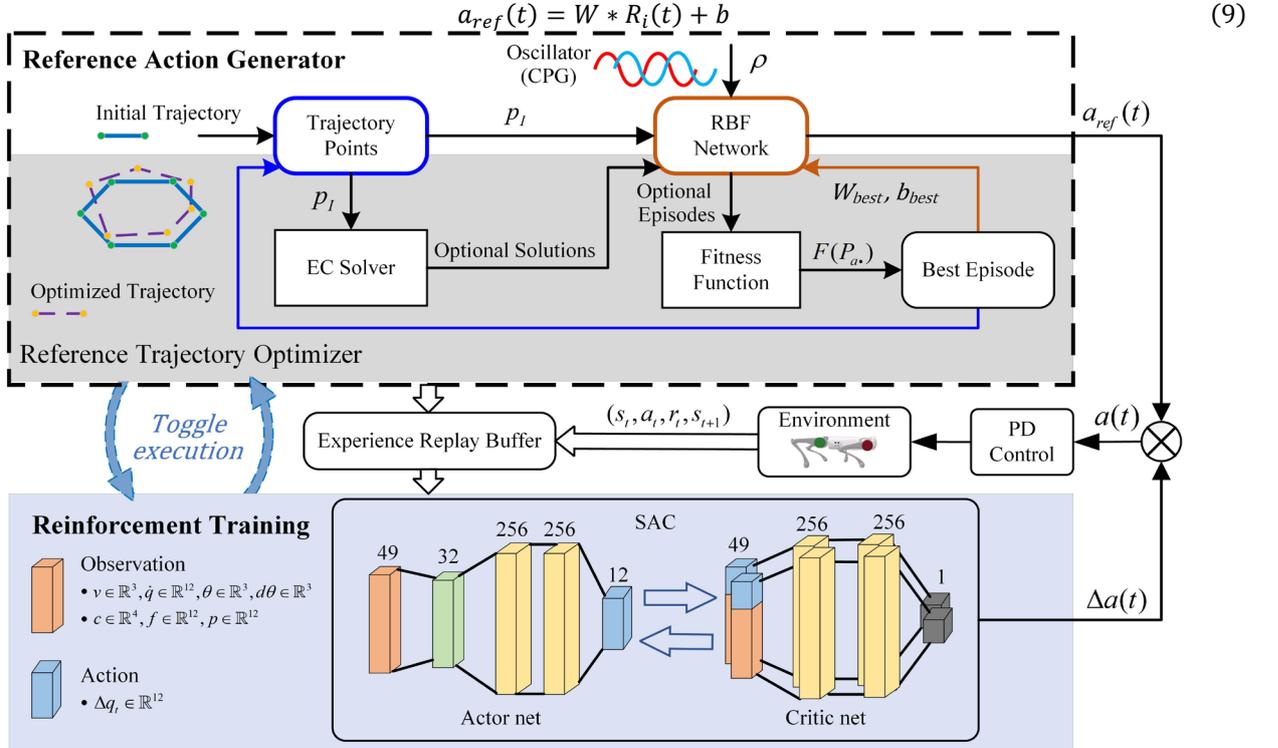

Fig. 2. A reinforcement training framework for walking gait generation of quadruped robots.

**Algorithm 1** Reference Action Generator
**Input:** $\varphi_0$, prior trajectory
  $k$, number of samped points
  $\delta$, threshold value
  $F$, fitness of the sampled points
  $N_{EC\_episode}$, train episodes for EC
**Output:** $\{W_{best}, b_{best}\}$, best weight and bias
1: Initialize RBFN parameters: $W, b \leftarrow W_0, b_0$
2: Initialize trajectory: $\varphi \leftarrow \varphi_0$
3: sample $k$ points from $\varphi$: $p_I$
4: $F_{best}$ = EC Evalute from Algorihtm 3
5: **for** i in $\{1...N_{EC\_episode}\}$ **do**
6:   sample $N$ groups of added points from EC solver: $P_a$
7:   **for** n in $\{1...N\}$ **do**
8:     update $p_I$: $p_I = p_I + p_a$
9:     computer distance: $\nabla_{\varphi_k} = \varphi - \varphi_k$
10:    **while** $\nabla_{\varphi_k} > \delta$ **do**
11:      update $W, b$
12:    **end while**
13:    evaluate the fitness of $p_a$: $F_i$ =EC Evalute from Algorihtm 3
14:    **if** $F_i > F_{best}$ **then**
15:      $F_{best} = F_i$
16:    **end if**
17:  **end for**
18: **end for**
19: take $W, b$ of $F_{best}$ as $W_{best}, b_{best}$
20: update RBFN parameters: $W, b \leftarrow W_{best}, b_{best}$

In Reference Trajectory Optimizer, a genetic algorithm derived from evolutionary computation is introduced to implement self-improvement of the reference action. It is worth noting that the genetic information used in traditional genetic algorithms is usually typical gait parameters such as step length, swing height, etc. These parameters may limit the search area of the action, and the interdependence between parameters makes it difficult to be optimized. Instead, we take a direct search for foot trajectory in the foot motion space, which is easily combined with foot trajectories generated by other arbitrary means (e.g., CPG) as movement priors. As shown in Fig. 3, the spatial position vectors (indicated by green circles) of several uniformly distributed green markers on the footpath trajectory are each used as genetic information that is adjusted based on a perturbation genetic algorithm that includes crossover or mutation process to enable these position vectors to be adjusted to arbitrary new positions (indicated by yellow circles). The trajectory consisting of these markers with new positions is then used to extend the search area of optimization, which has an untypical profile and is significantly different from the usual elliptical profile.

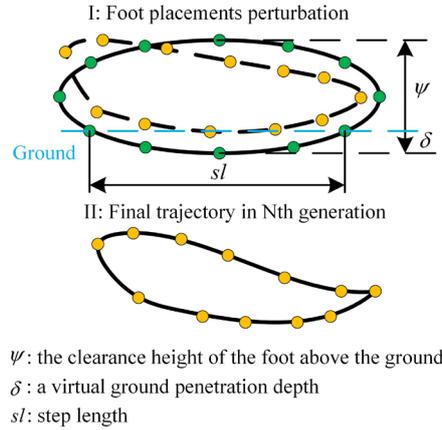

$\psi$: the clearance height of the foot above the ground
$\delta$: a virtual ground penetration depth
$sl$: step length

Fig. 3. The spatial profile of the foot trajectory is modified based on evolutionary computation.

Specifically, the initial foot trajectory $\varphi_0$ contains $k$ markers, and the position vector corresponding to each marker point is denoted as $P_I$. The set of $P_I$ is taken as the reference trajectory of the foot at that moment. Then the $k$ additional vectors $v_a$ with initial values of $[0,0]$ are taken as incremental values of $P_I$ to be used to update $P_I$ and therefore to update the reference trajectory. We use the Evolutionary Computation Solver to update the additional vector $v_a$. To ensure the breadth as well as the optimization of the exploration reference trajectory, the evolutionary computation solver updates $N_{\text{EC\_episode}}$ episodes each time $N_{\text{updateRAG}}$ steps is completed, generating $N$ new sets of $v_a$ as alternatives in each episode. The fitness criterion is used to evaluate the search effectiveness of each set of alternative $v_a$, i.e. to evaluate the performance of the corresponding alternative reference trajectory. The evaluation process is shown in Algorithm 2. The main process is to first sample the total reward in one episode (containing $N_{\text{EC\_step}}$ steps) walking evaluation and use them as the individual fitness $F_i$, then rank all the fitness to select the optimal $v_a$ and update the RBFN based on the corresponding $W_{best}$ and $b_{best}$.

---

**Algorithm 2** EC Evaluate

**Input:** $N_{EC\_step}$, train steps for EC in each episode
$\pi_{\theta_{fixed}}$, current fixed policy from RL
$\varepsilon_t$, random noise
$B$, experience replay buffer
$R_t$, status of RBFN neurons
$W, b$, RBFN weight and bias

**Output:** $fitness$

1: **for** i in $\{1...N_{EC\_step}\}$ **do**
2:     refence action: $a_{ref}(t) = W * R_t + b$
3:     action from RL: $\Delta a(t) = \pi_{\theta_{fixed}}(s) + \varepsilon_t$
4:     control signal: $a(t) = a_{ref}(t) + \Delta a(t)$
5:     execute action $a_t$ and observe reward $r_t$ and new state $s_{t+1}$
6:     fitness: $fitness = fitness + r_t$
7:     append $(s_t, a_t, r, s_{t+1})$ into $B$
8: **end for**

---

Reinforcement learning plays an incremental learning role in the proposed framework, and we specifically use the SAC algorithm responsible for training the quadruped robot to interact with its environment. The reference action $a_{ref}(t)$ generated by the Reference Action Generator is used as the reference action at the beginning of training, based on which the joint angle increment $\Delta a(t)$ is trained by reinforcement learning to further optimize the gait of the quadruped robot. It should be noted that both parameter updates in the Reference Action Generator and parameter updates in reinforcement learning result in different behavior distributions. Therefore, to make the training more stable, the parameters of the reinforcement learning policy are locked when the reference action is updated, and the parameters in the Reference Action Generator are locked when the parameters of the reinforcement learning policy are updated, and these two processes are executed alternately. Specifically, the parameters in Reference Action Generator are updated every $N_{updateRAG}$ steps, as detailed in Algorithm 3.

```
Algorithm 3 Reinforcement Training
Input:  B, experience replay buffer
        N_max steps, maximum train steps for RL
        N_updateRAG, update trajectory every N steps
        ε_t, random noise
        R_t, status of RBFN neurons
Output: Δa(t), quadruped action
1:  Initialize replay buffer B
2:  Initialize the policy and reference action generator
3:  for 0 ≤ steps ≤ N_max steps do
4:      Initialize observation s_0, action a_0, etc.
5:      if reach N_updateRAG then
6:          optimize the trajectory shape with Algorithm 1
7:      end if
8:      while env is not done do
9:          reference action from Algorithm 1: a_ref(t) = W_best * R_t + b_best
10:         action from RL: Δa(t) = π_θ(s) + ε_t
11:         control signal: a(t) = a_ref(t) + Δa(t)
12:         execute action a_t and observe reward r_t and new state s_{t+1}
13:         append (s_t, a_t, r_t, s_{t+1}) into B
14:         sample a random minibatch of transitions (s_t, a_t, r_t, s_{t+1}) from B
15:         for each gradient step do
16:             θ_i = θ_i - λ_Q ∇̂_{θ_i} J_Q(θ_i) for i ∈ {1,2}
17:             φ = φ - λ_π ∇̂_φ J_π(φ)
18:         end for
19:     end while
20: end for
```

**Observation Space.** The input state $s_t \in \mathbb{R}^{49}$ contains the three-dimensional linear velocity of the body $v = [v_x, v_y, v_z]$, the twelve-dimensional angular velocity of the joints $\dot{q} = [\dot{q}_0, \dot{q}_1, \dot{q}_2, \ldots, \dot{q}_{11}]$, the three-dimensional pose of the body $\psi = [R, P, Y]$, the differential of the three-dimensional pose of the body $\varphi = [dR, dP, dY]$, the four-dimensional foot touchdown state $c_f = [c_0, c_1, c_2, c_3]$, the twelve-dimensional foot contact force $F_f = [F_{x0}, F_{y0}, F_{z0}, \ldots, F_{x3}, F_{y3}, F_{z3}]$, and the twelve-dimensional foot position relative to the body $p_f = [p_{x0}, p_{y0}, p_{z0}, \ldots, p_{x3}, p_{y3}, p_{z3}]$.

**Action Space.** The action $a_t \in \mathbb{R}^{12}$ output from the reinforcement learning is the position increment $\Delta q_t$ of 12 joints, which is added to the reference action provided by the Reference Action Generator to obtain the expected value of the joint position at the current moment, as shown in Eq. (10).

$$\hat{q}_t = q_t^{ref} + \Delta q_t \tag{10}$$

The PD controller is used to calculate the joint torque τ, as shown in Eq. (11).

$$\tau_t = K_p(\hat{q}_t - q_t) + K_d(\hat{\dot{q}}_t - \dot{q}_t) \tag{11}$$

where $K_p$ and $K_d$ are kept constant in the simulation. The parameters $\hat{q}_t$ and $\hat{\dot{q}}_t$ represent the expected joint position and the desired joint velocity, respectively, and $\hat{\dot{q}}_t$ is set to 0.

**Reward function.** The reward function is designed to enable the robot to achieve desired target speed while successfully learning the gait, and to remain stable and energy efficient during movement. Therefore, the reward function is designed as $\omega_1 R_v + \omega_2 R_e + \omega_3 R_b + \omega_4 R_f + \omega_5 R_c + \omega_6 R_u$, where $\omega = [1.5, 0.07, 0.6, 0.3, 0.1, 0.1]$. The desired velocity of the body is defined as $V_d$, the actual velocity of the body is defined as $V_c$, τ is the torque of the motor, $\dot{q}$ is the angular velocity of the motor, $k$ is the number of motors, $V_{f,l}$ denotes the foot velocity when the $l$th foot touches the ground, $C_t$ denotes the number of current support legs, $C_d$ denotes the number of support legs in ideal condition, and $C_u$ denotes the number of parts where unexpected contact occurs. A curriculum factor $c_k = 1 - tanh(4.5 * min(V_x -

$V_d, 0)^2$) is introduced to make the robot prioritize learning the primary task (e.g., following the desired velocity) and to prevent the tendency to stand still [29]. As the speed of acquisition gradually reaches the target during training, $c_k$ increases accordingly to raise the weight of the other objectives in the total objective; the coefficients $c_b = 4$, $c_f = 2.5$ are set. Then the specific components of the award in each step are shown in Table 1.

Table 1. Reward components.

| Index | Award item | Symbol | Expression |
|---|---|---|---|
| 1 | Linear velocity of the base | $R_v$ | $R_v = min(V_d, V_c)$ |
| 2 | Energy consumption | $R_e$ | $R_e = -c_k \sum |(\tau * \dot{q}) * \Delta t|$ |
| 3 | Base motion | $R_b$ | $R_b = c_k(tanh(c_b||\omega_{xy}||^2) - 1)$ |
| 4 | Foot velocity variation | $R_f$ | $R_f = \sum_{l \in \{foot\ in\ contact\}} c_f min(V_d, V_c)$ |
| 5 | Foot touchdown condition | $R_c$ | $R_c = -max(C_t - C_d, 0)$ |
| 6 | Unexpected contact | $R_u$ | $R_u = -C_u$ |

The linear velocity reward $R_v$ is used to encourage the robot to track target velocity. The energy consumption reward $R_e$ is used both to penalize excessive joint torque to prevent damage to the joint actuators and to reduce the energy consumption of the motion. The base motion reward $R_b$ is used to penalize excessive rolling movements and excessive pitching movements to allow the body to remain stable during travel. The foot velocity variation reward $R_f$ is used to limit the foot velocity to prevent the robot from slipping and excessive contact force between the foot and the ground resulting in structural damage. Foot contact reward $R_c$ is a crucial reward for rapid learning of the target gait, and excessive foot contact is penalized to allow each leg to lift or touch the ground correctly according to the preset phase. The unexpected contact $R_u$ is used to penalize unreasonable collisions and contacts, such as the contact between the body and the ground.

**Experience Replay Buffer.** In the proposed framework, Reference Action Generator shares the same experience replay buffer with the reinforcement training process. Since evaluation of fitness in Reference Trajectory Optimizer is computed similarly to the total reward obtained in one episode for a new trajectory reinforcement learning-based walking training task, the sample-based experience generated from each fitness evaluation is stored in the experience replay buffer, thus providing more samples for learner updates in reinforcement learning. This data reuse mechanism allows experience to be transferred from the evolutionary population to the reinforcement learning learners, providing experience that may contribute to higher long-term reward while improving sampling efficiency [30].

## 4. Results

### 4.1. Test Conditions

Pybullet software, with version 3.2.0, is used as the simulation platform for this study. The quadruped robot used in the simulation is the GO1 quadruped robot, which contains three independent joints in each leg, thus enabling the movement of each leg and body in 3D space. The position angle and angular velocity of each joint can be measured directly, body posture and locomotion acceleration can be measured by IMU, and the contact force and contact state between each foot and the ground can be measured by the force sensor on the foot. The parameters of the robot are shown in Table 2.

Table 2. Robot parameters.

| Parameter | Value |
|---|---|
| Base mass / Leg mass | 10.5 kg / 2 kg |
| Number of joints | 3*4 |
| Max motor torque | 33.5 N*m |
| Hip length / Thigh length / Shank length | 0.1 m / 0.2 m / 0.2 m |
| Abd kp/kd | 80 / 2 |
| Hip kp/kd | 120 / 4 |
| Knee kp/kd | 90 / 3 |
| Initial position | [0, 0, 0.26 m] |
| Initial motor angles | [0, 0.9 rad, -1.8 rad] |

The primary training goal is to make the quadruped robot walk stably in regular terrain, with expected speed and gait. Therefore, we design varied terrains for simulation, including basic terrains such as horizontal ground, slope, and stairs, as well as several combined terrains such as continuous up-downslope and up-downstairs, which are depicted in Fig. 4.

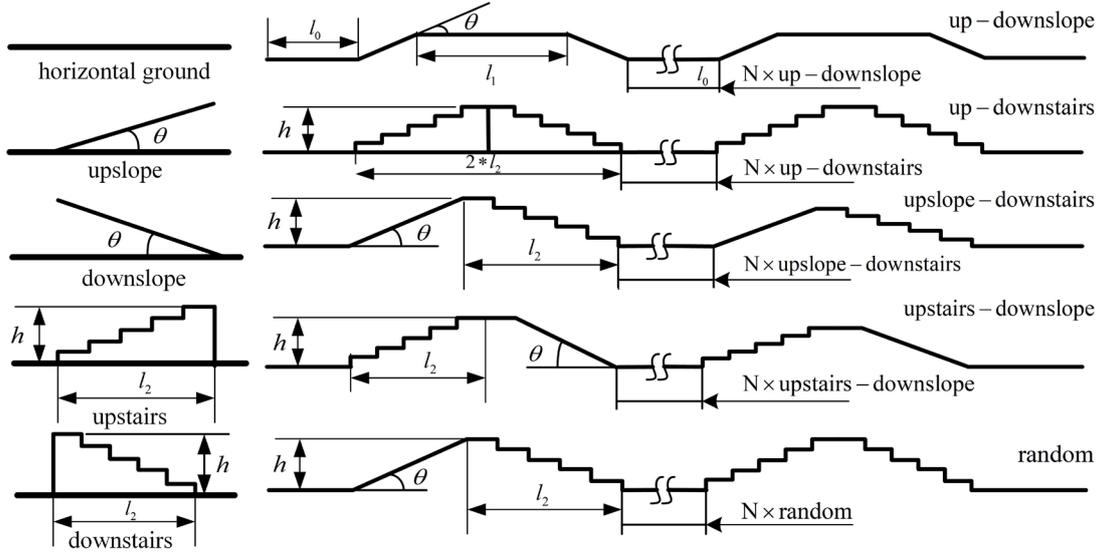

Fig. 4. Base and combined terrains used in training simulation. (a) Simple base terrain. (b) Combined terrains that are built from direct combination of different base terrains. Ten same combined terrain units are combined again directly for testing, i.e., N=10. Other parameters used to create terrain include $l_0 = 1$ m, $l_1 = 1$ m, $l_2 = 1.65$ m, $h = 0.15$ m.

The total number of steps for each training task is set to one million, and several groups are employed to distinguish the respective goal and characteristic of these training tasks. The specific description of these groups is shown in Table 3.

Table 3. The groups used in the test.

| Group | RL algorithm | Dimensions of state space | | Reference Trajectory | | |
|---|---|---|---|---|---|---|
| | | 37 | 49 | Fixed | Optimized | Optimization method |
| Group 0 | SAC | √ | | √ | | - |
| Group 1 | SAC | | √ | √ | | - |
| Group 2 | SAC | | √ | | √ | Genetic algorithm |
| Group 3 | SAC | | √ | | √ | Uniform distribution |
| Group 4 | SAC | | √ | | √ | Normal distribution |
| Group 5 | SAC | √ | | | √ | Genetic algorithm |

The parameters of the SAC algorithm and the neural network used in the tests are shown in Table 4.

Table 4. SAC hyperparameters and neural network size.

| Parameter | Symbol | Value |
|---|---|---|
| batch size | | 256 |
| initial steps | | 10000 |
| replay buffer size | | $10^6$ |
| learning rate | $lr$ | $3 \times 10^{-4}$ |
| temperature parameter | $\alpha$ | 0.2 |
| discounted factor | $\gamma$ | 0.99 |
| number of hidden layers | | 2 |
| number of hidden units per layer | | 256 |
| activation | | tanh |

The results and analysis of the tests by different factors are as follows.

## 4.2. Observation Space

The dimension of the observation space can have an impact on the reinforcement learning training; it is generally believed that too few observation dimensions make the description of the current state inaccurate or missing, and too many observation dimensions can lead to curse of dimensionality. We first investigate the effect of increasing only the state space dimension on the motion performance of the quadruped robot and the improvement of terrain adaptation from the aspect of reinforcement learning. For this purpose, we consider two observation spaces with different proprioception: full observation and partial observation. Since training an inexperienced robot is a very difficult task, for this reason we provide a pre-defined foot trajectory as a reference. It is worth noting that the reference trajectory used in the test are roughly given and not necessarily the appropriate reference.

**Full observation (49 dim)**: Full observation focuses on perception of both movement speed and proprioception, specifically body speed $v$ (body state), joint angular velocity $\dot{q}$ (joint state), foot contact state $c_f$, foot contact force $F_f$, and the relative position of the foot $p_f$.

**Partial observation (37 dim)**: Partial observation removes only the relative position of the foot on the basis of the full observation.

The rewards obtained from training based on each of the two observation spaces in different terrains that as shown in Fig. 5(a) are shown in Fig. 5(b), where the blue line corresponds to Group 0 and the red line corresponds to Group 1. In the tests on all terrains, Group 0 achieves the training goal only on horizontal ground, while by introducing additional information $p_f$, which is closely related to gait movement, Group 1 obtains an increase of about 67% in average reward value on horizontal ground compared to Group 0 and obtains a higher reward value on 15° slope. In addition, the reward of Group 1 in 15° slope shows significant fluctuation, indirectly reflecting that Group 1 has some chance to explore the proper policy, which indicates that increasing the observation dimension by adding foot position information has an enhanced effect on training in simple terrain. In contrast, for steeper slope or more complex terrain, the richness of the observation dimension does not lead to a significant improvement in the training effect. Considering that the reference trajectory used in these two groups of tests is arbitrarily set, we suppose that even a suitable design of observation dimension and reward function may not achieve the desired training effect if it is based on an unreasonable reference trajectory. This implies that improving the initial reference trajectory rather than adopting a constant reference trajectory is necessary for gait learning in quadruped robots.

### 4.3. Reference Trajectory Optimization

Since a greater variety of observation dimensions more or less improves the training effect, we directly test the gait training framework based on full observations. In this test, each time the reference trajectory is updated, the framework automatically adjusts the reference trajectory by using a genetic algorithm to search for 10 episodes, generating 40 sets of candidate solutions in each episode, and then selecting the unique optimal solution among the 10 episodes as the result, and using the result as the reference trajectory for incremental learning thereafter. To avoid the lack of sample richness at the beginning of the training, 10,000 steps of training is performed using reinforcement learning before the reference trajectory is updated for the first time and the corresponding result data is stored in the Experience Replay Buffer. From the training results corresponding to Group 0 and Group 1, it can be inferred that the initial reference trajectory set by experience is not entirely suitable, so the reference trajectory is updated by optimization for the first time when the total number of training steps reaches 10,000, and then updated again every 50,000 steps.

The training results that obtained after including the reference trajectory update mechanism are shown in Fig. 5(c), where the green line corresponds to Group 2. It can be noticed that Group 2 has significantly improved the training effect on each terrain compared with Group 1. In the simplest horizontal ground test, the average reward value obtained by Group 2 increased by about 147% compared to Group 1, while for all other training tests on more complex terrain, Group 2 is successful and obtains higher rewards. Further, based on the analysis of the reward composition, it can be found that the reward values of Group 2 are all higher than those of Group 1. This result fully illustrates that the policy learned through the introduction of the reference trajectory improvement mechanism is more likely to find better combination patterns under multimodal rewards, and can effectively improve the adaptability of the quadruped robot to complex terrain.

### 4.4. Trajectory Update Rule

Training tests based on Group 3 and Group 4 are performed and the test results are compared with the results based on Group 2 to prove that the trajectory optimization based on genetic algorithm search is reasonable. Group 3 takes a uniform distribution criterion to randomly sample additional vectors in the range $[0, 0.01]$, while Group 4 uses a standard normal distribution criterion ($\mu = 0$, $\sigma = 0.01$) for random sampling. Two regular terrains, horizontal ground and up-downslope, are used for the test, where three groups have exactly the same trajectory update frequency, number of update episodes, number of candidate solutions per episode and fitness evaluation criteria. The test results are shown in Fig. 5(f), where the yellow line represents Group 3 and the pink line represents Group 4.

The results of the fitness evaluation show that each of the three different update methods for the additional vectors explored the correct policy in the horizontal ground test, and it is noted that the trajectory updated by the genetic algorithm based on the genetic algorithm shows a higher fitness compared to the other two methods. In the up-downslope tests, the advantage of the genetic algorithm is more obvious, as both Group 3 and Group 4 are unable to explore a suitable reference trajectory during the whole training process, thus leading to the failure of the training task, while Group 2 is quickly updated to a suitable reference trajectory and completes the training task. In addition, the sampling ranges of the uniform and normal distributions are artificially set, and we find that the probability of failure is very high if the sampling range is set large, so the regulation of the parameters of these two sampling methods needs to rely on expert experience. This shows that if a random reference trajectory is used for exploration, or if a reference trajectory distributed within a region based on some rules is used for exploration, it is not as effective as a reference trajectory based on genetic algorithm.

### 4.5. Suitability of Reference Trajectory Optimization

Considering that the proposed gait training framework shows a better improvement compared to the conventional reinforcement learning framework in the full observation dimension test, here we continue to test this framework based on partial observation dimension, and the experimental results are shown in Fig. 5(d). It can be found that the training success rate corresponding to Group 5 is significantly higher compared to the results of Group 0, which indicates that the reference trajectory optimization mechanism is generally effective for enhancing reinforcement learning-based gait training.

### 4.6. Walking Performance

After the training is completed, the performance of the policy in quadruped robot walking is then analyzed based on specific factors such as power consumption, walking balance, and anti-disturbance during movement. The Wide Stability Margin (WSM) is used to evaluate the walking stability, which is defined as the distance $d$ from the projection point of the robot's center of gravity to the intersection of the two diagonals of the support polygon. The smaller the value of $d$, the better the stability. In addition, the total power $p_w$ of the robot is defined as the sum of the product of the torque of each joint actuator and its angular velocity, as shown in Eq. (12).

$$p_w = \sum_{i=1}^{m} \sum_{j=1}^{n} |\tau_{ij} \dot{q}_{ij}| \tag{12}$$

where $m$ is the number of legs and $n$ is the number of joints in each leg.

Horizontal ground and combined terrain are used to test the locomotion performance of the quadruped robot. The performance corresponding to the best policy obtained by training on horizontal ground is shown in Fig. 6(a) to (g), where all the data are taken from a period of time after starting the movement from the standing state. After the robot completes the transition from the standing state to the walking state within 2 seconds (corresponding to the first 100 training steps), the angles of all joints start to show a stable periodicity, as shown in Fig. 6(a), while the posture of the body and the height of the center of mass are kept to change within a certain range, and the quadruped robot keeps walking stably. Fig. 6(d) shows the optimization process of the foot trajectory during the training process. Since the genetic algorithm selects the optimal solution to update the foot position $P_I$ after a multi-round search of the additional vectors, each change in the spatial profile of the foot trajectory is considered to be more optimal compared to the one before the update. The optimal foot trajectory profile shown in Fig. 6(d) is used as the reference trajectory, and the foot trajectory obtained after incremental reinforcement learning based on this reference trajectory is shown in Fig. 6(e). In addition, conditions under different desired walking speeds (0.5m/s, 1.5m/s, 2.5m/s) and random disturbance are tested, and Fig. 6(b) shows the body state at different speeds, and the disturbance is imposed as shown in Fig. 6(h). For higher desired speed, the trained policy can still keep the variation of pitch angle, roll angle and center-of-mass height of the body within a certain range. Moreover, the motion state can be recovered quickly after imposing random disturbance. If the desired speed is set to 0.5m/s, 1.5m/s and 2.5m/s, the average speed after training is 0.59m/s, 1.64m/s and 2.49m/s, respectively, as shown in Fig. 6(c). Further, the test results of power consumption and WSM for different desired speed conditions are shown in Fig. 6(f) and Fig. 6(g), respectively. It can be observed that as the target speed increases, the power consumption will increase accordingly and the stability will decrease. It is worth noting that the policy trained by exerting random perturbations during training has insignificant changes in power consumption, but has better walking stability.

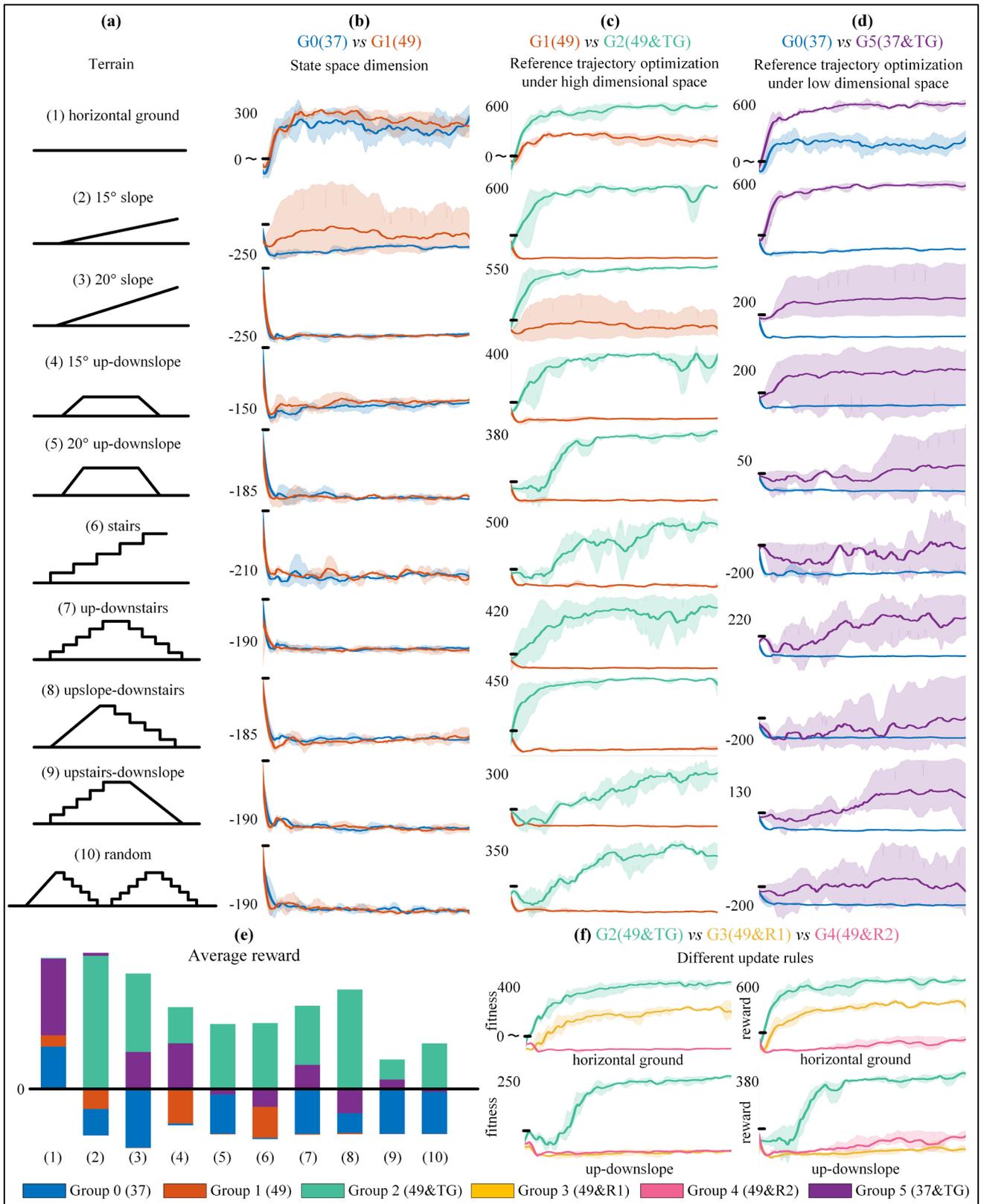

Fig. 5. Rewards for training in a variety of terrains. (a) Terrains for testing. (b) Comparison of training based on different observation dimension, i.e., Group 0 vs Group 1. (c) Comparison of training based on whether to refer to the trajectory optimization mechanism, i.e., group 1 vs group 2. (d) Comparison of training based on whether to refer to the trajectory optimization mechanism in lower dimension condition, i.e., group 0 vs. group 5. (e) Comparison of average rewards obtained by Group 0, 1, 2, and 5. (f) Comparison of training based on different reference trajectory update methods, i.e., comparison between groups 2, 3, and 4.

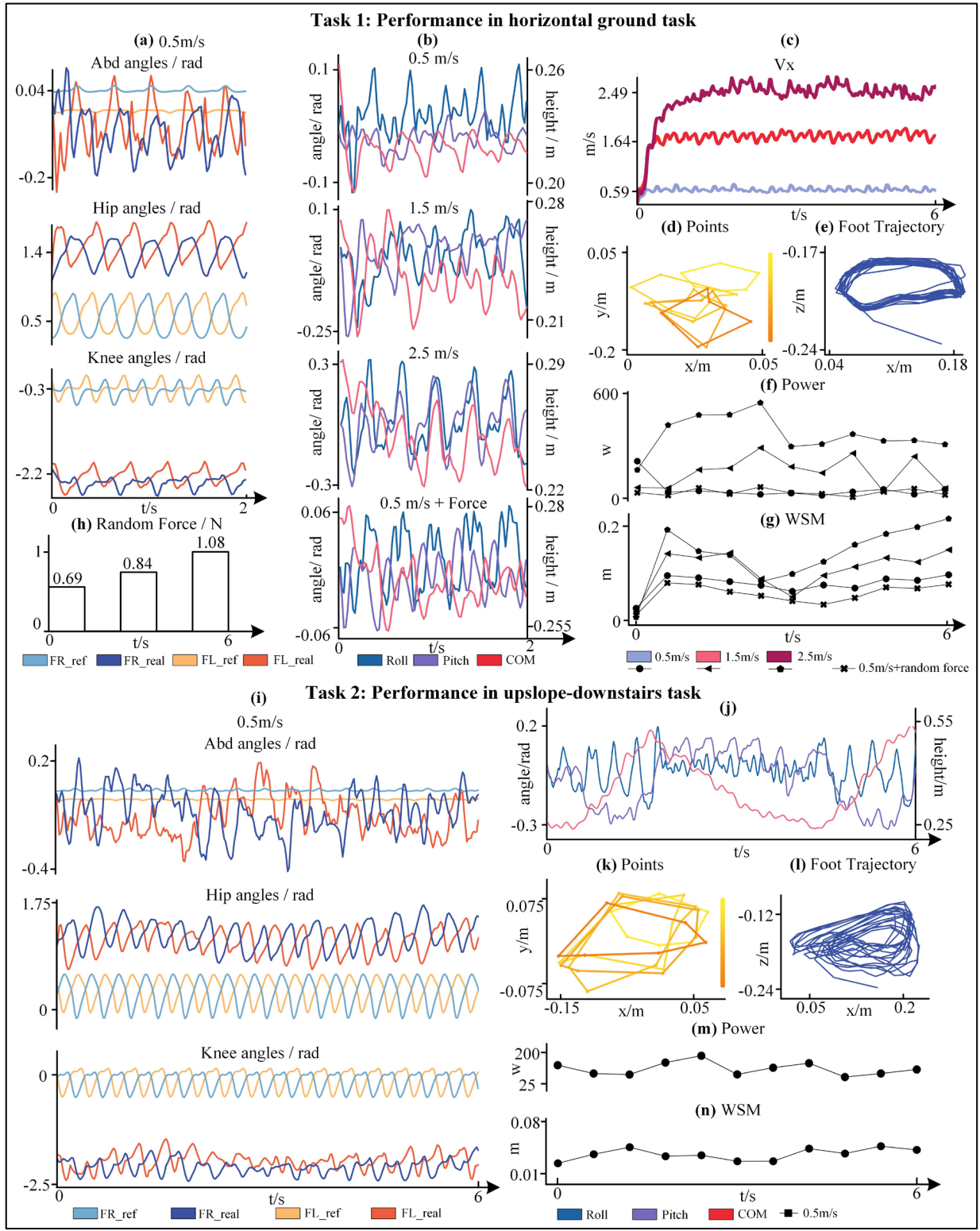

Fig. 6. Evaluation of the locomotion performance of the gait policy acquired from training, (a) to (h) correspond to training tests on horizontal terrain, and (i) to (n) correspond to training tests on upslope-downstairs terrain. (a) Optimal reference movements (light lines) for both front legs within 2 seconds (100 training steps) and actual movements (dark lines) after training. (b) The roll angle, pitch angle and height of the CoM at different speeds, where the last is when an external force is exerted. (c) The actual average speed when the desired speed is 0.5m/s, 1.5m/s and 2.5m/s respectively. (d) The evolution of the foot reference trajectory within 1 million steps, which is obtained based on genetic algorithm search and corresponding to a change in color from light to dark. (e) Foot trajectory of the quadruped robot under optimal policy. (f) Power consumption in 6s (300 training steps) with different speeds and disturbance. (g) WSN in 6s with different speeds and disturbance. (h) External force used to simulate random disturbance is applied at 1s intervals, each lasting 1s. (i) Optimal reference movement of the two front legs and the actual movement in the test. (j) The roll angle, pitch angle and height of the CoM at a speed of 0.5m/s. (k) to (n) are similar to (a) to (g), and the difference is mainly in terrain.

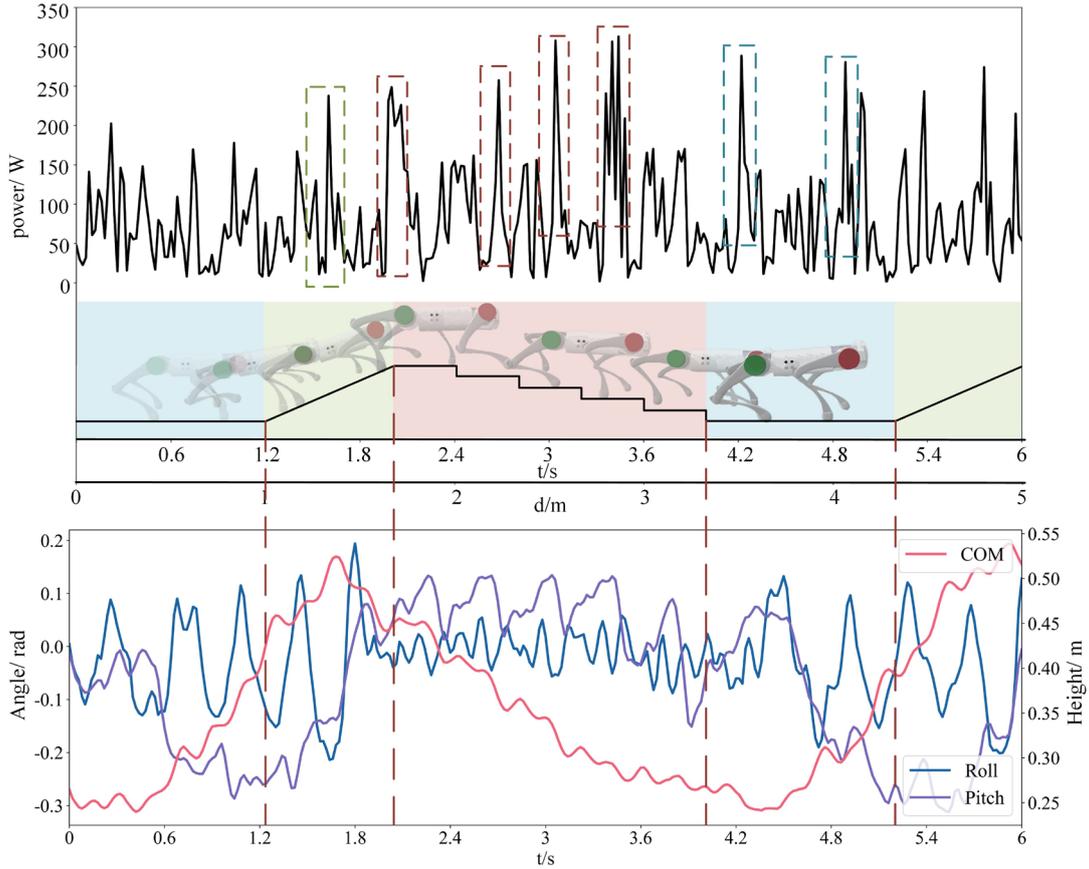

Fig. 7. Detailed analysis of upward slope and downward stairs.

The performance of the quadruped robot in 6 seconds (corresponding to the first 300 training steps) during the upslope-downstairs is shown in Fig. 6(i) to Fig. 6(n). The results show that the trained policy enables the quadruped robot to maintain a stable rhythmic walking even in the combined terrain, as shown in Fig. 6(i) and Fig. 6(j). The reference trajectory of the foot is optimized over a larger area, as shown in Fig. 6(k) and Fig. 6(l). In addition, the power consumption and WSM are kept within a certain range, as shown in Fig. 6(m) and Fig. 6(n). The detailed process of walking in this terrain is shown in Fig. 7, where the power is the instantaneous value and not the average value as shown in Fig. 6(m). The sudden change in power consumption first occurs during the upward slope period of about 0.6 to 1.8 seconds, when the quadruped robot uses more strength to adjust its posture to maintain balance and thus adapt to the transition from flat to slope. When going uphill, the hip and knee joints of the front legs are bent, the hip and knee joints of the rear legs are extended, and the pitching motion of the body changes greatly. Another significant change occurs during the slope-to-stairs transition from 1.8 to 2.4 seconds, with corresponding changes in body pitch and roll movements, and the change in posture well explains why the WSM becomes larger. In the process of going down the stairs from 2.4 to 4.2 seconds, the pitch angle shows obvious periodic changes, and the power changes are also significant because the quadruped robot adjusts its posture to maintain balance by increasing the motion amplitude of each joint in this process. Moreover, since the most critical training objective is stable walking rather than energy conservation, the reward term on energy has a low weight in the total reward function, which is an important reason for the apparent power increase condition during down stairs or during terrain transition.

The power consumption and WSM of the quadruped robot during walking on different types of terrain are shown in Fig. 8 and Fig. 9, respectively, for a time range of 6 seconds from the start of movement, which corresponds to 300 training test steps. In these tests, the power consumption of the robot basically remained within a certain range, and only during the process of going up the stairs the power consumption becomes significantly larger, which includes both the power demand of rapidly lifting the body and also due to the process of transitioning from the flat ground to the stairs, the robot needs more strength to adjust the body posture to maintain balance. Accordingly, the WSM during upstairs is also the largest.

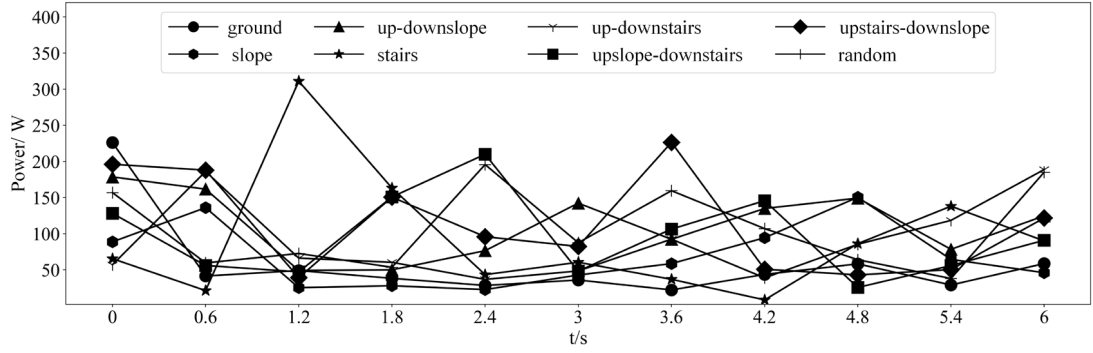

Fig. 8. Comparison of power consumption of the quadruped robot during walking on different terrain.

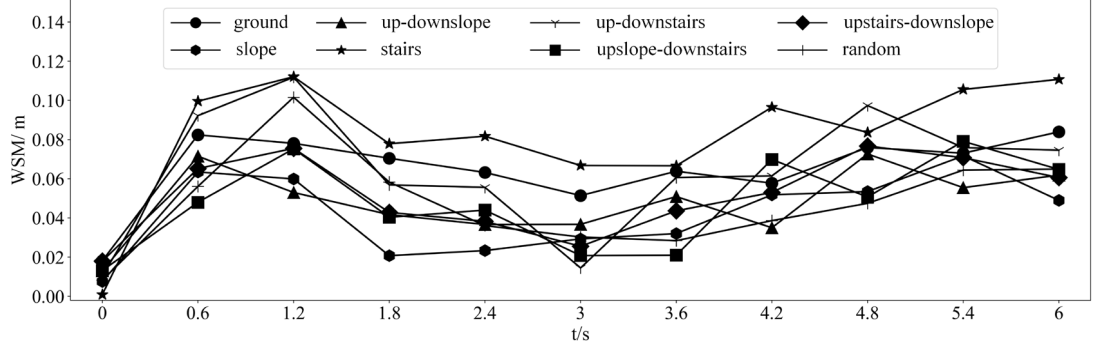

Fig. 9. Comparison of WSM of the quadruped robot during walking on different terrain.

### 4.7. Parallel Training

The framework proposed in this paper enables parallel training to reduce the time cost of the training process, and the scheme we design by incorporating the parallel mechanism into the reinforcement training component is shown in Fig. 10. In this scheme, the Reference Action Optimizer and Global Network are executed alternately on a main thread running on CPU 0, as shown in the purple area. Multiple parallel Agents responsible for reinforcement learning are built independently during training and interact with independent environment to collect independent experience. We use Parl to distribute all the Agents into separate CPUs, as shown in the blue area. It should be noted that simultaneous update is introduced, i.e., in each training episode, the Global Network waits for each Agent to complete the current episode individually, then aggregates and averages the gradient data uploaded by these Agents to obtain a uniform gradient and updates the parameters of the main network with it, and finally updates all the Agents simultaneously using these parameters. The computing platform used in this paper is based on the Intel(R) Core™ i9-10900X CPU @ 3.70GHz, and 64G of RAM. Based on this hardware configuration, we test the parallel framework in two types of terrain, horizontal ground and upslope-downstairs, which consume 8h35m and 9h52m in single process test, and 4h26m and 4h19m when using 10 Agents for parallel computation, this result shows that the elapsed time for parallel training is reduced by about 50% compared to the non-parallel one. If the Reference Action Generator is also designed in parallel, the training time can be further reduced.

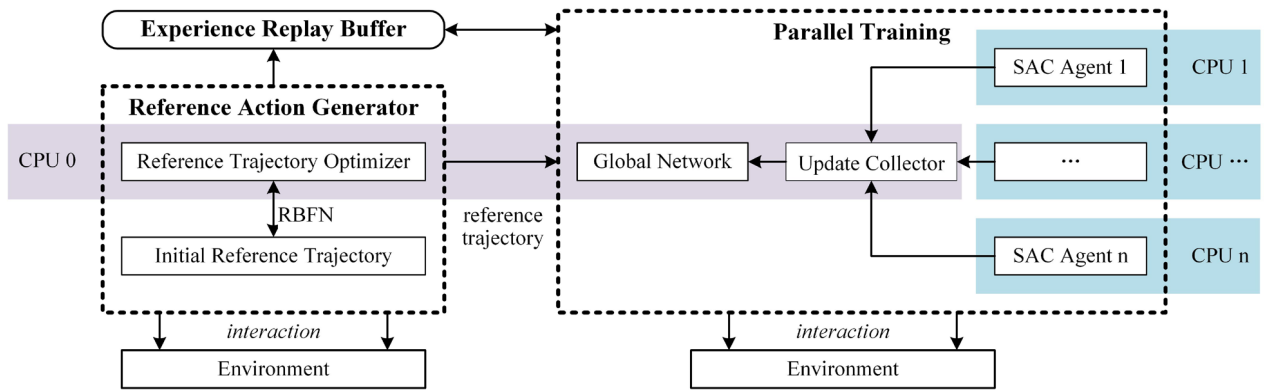

Fig. 10. Parallelized training framework.

# 5. Discussion

The constituents of the rewards obtained during training in two different terrain conditions, respectively, are shown in Fig. 11. It can be found that in the early stage of training, the undesired contact rewards $R_u$ account for the largest proportion of the total reward, which indicates that at this stage the robot is mainly in a fall or self-collision state and cannot walk normally. After training convergence, the largest proportion of speed rewards $R_v$ and basal movement rewards $R_b$ are observed. This indicates that as the reference trajectory is continuously improved and incremental learning continues, the robot gradually establishes a target policy that allows it to achieve the desired tracking speed while maintaining the stability of the body. During the training convergence process, the continued optimization of the reference trajectory still has an impact on the reward distribution. For example, in training on horizontal ground as shown on the left chart of Fig. 11, the situation where the reference trajectory is updated at the 100,000th step to obtain a higher walking speed at the cost of a larger motion bump (corresponding to a smaller $R_v$ and a larger $R_b$) is re-optimized after a period of exploration (corresponding to increasing $R_v$ and decreasing $R_b$), and the ratio of both $R_c$ and $R_e$ also changes significantly in this process, all of which indicate that the optimization of the reference trajectory after training convergence can continue to motivate the target policy to explore better combination patterns under multimodal rewards. In addition, the proportion of foot contact rewards $R_c$ and energy consumption rewards $R_e$ decreases, which also shows that gait is correctly learned while energy consumption is further reduced.

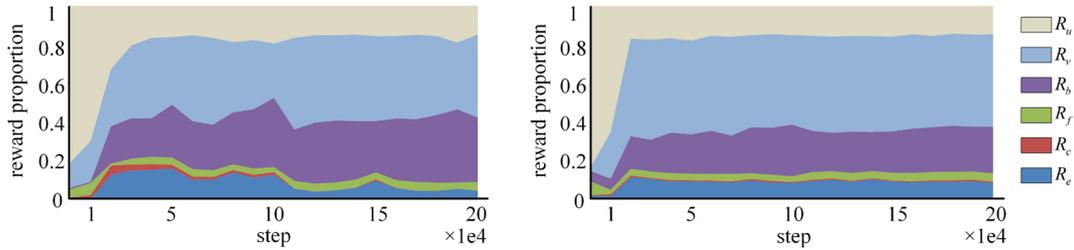

Fig. 11. Constituents of the rewards in training. The left chart corresponds to the training task on horizontal terrain, and the right chart corresponds to the task on upslope-downstairs terrain. The data for both charts are taken from the first 200,000 training steps of each.

The test results have sufficiently demonstrated that both enhancing the state space dimension and optimizing the reference trajectory are beneficial in improving the training success rate, but further observation of Fig. 5(e) clearly shows that optimizing the reference trajectory can be more effective in adapting to various terrain than simply increasing the state dimension. This may be due to the fact that adding specific state dimensions is only applicable to particular training tasks, and the reasonableness of the initially provided reference trajectory greatly affects the training effect, both of which make the benefits of simply adding state dimensions often not outstanding. In contrast, the reference trajectory can be improved by using the genetic algorithm to spontaneously perform optimal search for different objectives, and then gradually eliminate the unreasonable factors in the initial trajectory during the reference trajectory optimization to obtain more training gains, as well demonstrated by the tests in 15° and 20° slope.

Although the gait training framework proposed in this paper has obvious enhancement effects for both state spaces with different dimensions, the optimized reference trajectory may be searched faster based on the rich state dimension, and in addition setting a more appropriate initial reference trajectory can also improve the training efficiency.

In addition, to validate the selection of reinforcement learning algorithm, two other popular deterministic policy algorithms, DDPG and TD3, are tested and their results in the horizontal ground training are shown in Table 5. The results show that the maximum entropy-based stochastic policy performs better than the deterministic policy for the proposed approach.

Table 5. Average reward of different RL algorithms.

| Algorithm | Reward |
|---|---|
| DDPG | 291.6 |
| TD3 | 7.11 |
| SAC | 559.15 |

## 6. Conclusion

Reinforcement learning methods have been widely used to study gait generation in quadruped robots, not only because of their appealing random exploration and reward mechanism, but also because they are so similar to the process of learning to walk in humans and other animals - by continuously trying in order to eventually succeed. Although this romantic approach is part of many researchers' exploration of the nature of gait learning, it doesn't go entirely smoothly due to the differences between robots and animals in various aspects. This paper reconsiders the incremental reinforcement learning approach that has been commonly used from a behavioral bionic perspective, and identifies one of the important reasons for its staged success as the inheritance of progressive features in the evolution of motion behavior in animals. On this basis, we introduce another key factor in the evolution of motion behavior in animals, that is, a self-improvement mechanism for the reference gait. If the reference gait is obtained from real animal data, it further reflects the orientation of factors such as balance, energy consumption, and habituation that are embedded in the long process of acquiring this behavior in animals. Based on a more comprehensive imitation of the evolution of animal motion behavior, we construct a new framework for gait training to incorporate both incremental reinforcement learning of action and self-improvement of reference trajectory.

For the two main components of the framework, the genetic algorithm and the SAC algorithm are adopted to complete the final design of the framework, and the reasons for their selection are explained separately. Simulation tests in various terrains are performed in detail, all data results are recorded based on statistics and three main conclusions are obtained. First, compared with conventional reinforcement learning algorithms, the proposed framework achieves good results in all terrains under test; second, the self-improvement of the reference trajectory is the most significant in terms of its contribution to successful training, in contrast to the fact that simply increasing the state-space dimension is often futile in training tasks on complex terrains; third, the gait acquired by training with the proposed framework maintains a good level of stability and power consumption. Therefore, the research in this paper allows the application of reinforcement learning to gait training and even real quadruped robot traveling tasks to be advanced. In addition, this framework is partially parallelized to reduce the training time by about half, which is the preliminary work to proceed with the deployment of the framework in real robots, and prototype experiments will be carried out in the future.

## Acknowledgment

This work was supported by the National Natural Science Foundation of China [Grant number 61733002].